\documentclass[conference]{IEEEtran}
\IEEEoverridecommandlockouts

\usepackage{amsmath,amssymb,amsfonts}
\usepackage{array}
\usepackage{algorithmic}
\usepackage{graphicx}
\usepackage{textcomp}
\usepackage{xcolor}
\def\BibTeX{{\rm B\kern-.05em{\sc i\kern-.025em b}\kern-.08em
    T\kern-.1667em\lower.7ex\hbox{E}\kern-.125emX}}
    
\graphicspath{{Figures/PNG/}}
\usepackage{orcidlink}
\usepackage{bm,bbm}
\usepackage[numbers,compress]{natbib}
\usepackage{booktabs}
\usepackage{multirow}

\begin{document}

\title{\uppercase{LC4-DViT: Land-cover Creation for Land-cover Classification with Deformable Vision Transformer}}

\author{%
\IEEEauthorblockN{%
Kai Wang$^{1,2,*}$\orcidlink{0009-0000-4751-9553},
Siyi Chen$^{1,3,*}$\orcidlink{0009-0003-5929-4285},
Weicong Pang$^{1,4,*}$\orcidlink{0009-0007-8208-6723},
Chenchen Zhang$^{1}$\orcidlink{0009-0005-7446-1621},\\
Renjun Gao$^{5}$\orcidlink{0009-0005-0028-428X},
Ziru Chen$^{1}$\orcidlink{0009-0009-1732-3651},
Cheng Li$^{1,\dagger}$\orcidlink{0009-0009-4322-7591},
Dasa Gu$^{1,\dagger}$\orcidlink{0000-0002-5663-1675},\\
Rui Huang$^{2,\dagger}$\orcidlink{0000-0002-7950-1662},
Alexis Kai Hon Lau$^{1}$\orcidlink{0000-0003-3802-828X}
}
\IEEEauthorblockA{%
$^{1}$HKUST
$^{2}$SSE, The Chinese University of Hong Kong (Shenzhen)
$^{3}$JHU
$^{4}$NUS
$^{5}$MUST\\[0.2em]
$^{*}$Equal contribution.\quad
$^{\dagger}$Corresponding authors: Cheng Li (clieo@connect.ust.hk), Dasa Gu, Rui Huang.}
}

\maketitle

\begin{abstract}
Land cover underpins ecosystem services, hydrologic regulation, disaster-risk reduction, and evidence-based land planning, making timely and accurate land-cover maps essential for environmental stewardship. However, remote sensing-based land-cover classification is constrained by scarce and imbalanced annotations and by geometric distortions in high-resolution scenes. We propose LC4-DViT (Land-cover Creation for Land-cover Classification with Deformable Vision Transformer), which integrates description-driven generative data creation with a deformation-aware Vision Transformer. A text-guided diffusion pipeline leverages GPT-4o-generated scene descriptions and super-resolved exemplars to synthesize class-balanced training images, while DViT combines a DCNv4 deformable convolutional backbone with a ViT encoder to capture fine-scale geometry and global context. On eight AID classes (Beach, Bridge, Desert, Forest, Mountain, Pond, Port, River), DViT achieves 0.9572 OA, 0.9576 macro F1, and 0.9510 Kappa, improving over a vanilla ViT (0.9274 OA, 0.9300 macro F1, 0.9169 Kappa) and outperforming ResNet50, MobileNetV2, and FlashInternImage. Cross-dataset evaluation on a three-class SIRI-WHU subset (Harbor, Pond, River) yields 0.9333 OA, 0.9316 macro F1, and 0.8989 Kappa, indicating good transferability. An LLM-based judge (GPT-4o) that scores cosine similarity heatmaps further suggests DViT’s attention aligns best with hydrologically meaningful structures. Overall, combining description-driven generative augmentation with deformation-aware transformers is promising for high-resolution land-cover mapping. Project resources are available at  \href{https://charlescsyyy.github.io/LC4-DViT/}{https://charlescsyyy.github.io/LC4-DViT/}.
\end{abstract}

\begin{IEEEkeywords}
Remote sensing, image diffusion, land-cover classification, deformable convolution network, vision transformer.
\end{IEEEkeywords}


\begingroup
\renewcommand\thefootnote{}
\footnotetext{The original AID dataset and the SIRI-WHU dataset are publicly available at \href{https://captain-whu.github.io/AID/}{AID} and \href{https://opendatalab.com/OpenDataLab/SIRI-WHU/tree/main}{SIRI-WHU}, respectively. The work described in this paper was substantially supported by the grants from the Research Grants Council of the Hong Kong SAR, China (Project No. T24-614/25-N, C6001-24Y). The authors thank HKUST Fok Ying Tung Research Institute and the National Supercomputing Center in Guangzhou Nansha Sub-center for providing high-performance computational resources.}
\endgroup


\section{Introduction}
Land-cover is a foundational descriptor of the Earth's surface that shapes ecosystem services, biodiversity, carbon storage, hydrologic regulation, food security, and urbansustainability~\citep{Musetsho2022Ecosystem,Chen2024Water,li2023wetlands,allain2012wetland,Zhou2024Cooling}. Accurate, timely land-cover maps are therefore essential for environmental monitoring, spatial planning, and evidence-based policy, especially in rapidly changing urban and agricultural regions. Remote sensing-based land-cover classification offers a scalable path to such maps and has advanced rapidly with deep learning~\citep{ke2015ann,gui2022infrared}; yet deployment remains hampered by scarce and imbalanced annotations and by the difficulty of modeling narrow, deformation-rich landforms in high-resolution imagery, such as meandering rivers, small ponds, ports, and bridges~\citep{Mashala2023Systematic,Lupa2024Rivers,Li2023CA,cui2025river}.

Diffusion-based generative models have enabled high-fidelity image and video synthesis and powerful text-to-image systems~\citep{Ho2020DDPM,Song2020SDE,Nichol2021GLIDE,saharia2022photorealistic}. Their adaptation to remote sensing is more recent: RSDiff~\citep{Sebaq2023RSDiff}, SatDM~\citep{Baghirli2023SatDM}, DiffusionSat~\citep{Khanna2023DiffusionSat}, and CRS-Diff~\citep{Tang2024CRSDiff} synthesize auxiliary satellite scenes using text prompts, semantic masks, and metadata. However, most pipelines still rely on generic prompts and heuristic filtering~\citep{DeVita2024DMGS}, often yielding semantic inconsistencies and geometric artifacts that limit their value for high-precision land-cover mapping.

In parallel, land-cover mapping has progressed from pixel-wise statistical classifiers to deep-learning pipelines~\citep{Marmanis2016DeepLearning,cui2024gf2river,Tzepkenlis2023DeepSegmentation}. Convolutional neural networks (CNNs) fine-tuned from natural-image backbones dominate many benchmarks~\citep{Vali2020DeepLearning,Khan2022RemoteSensing,Zhu2019CNNs,Li2024GCTInception,Li2024WheatPest,Xia2017AID}, but their fixed receptive fields struggle with elongated shorelines, meandering rivers, and fragmented urban parcels~\citep{Khan2024TransformerBased,wang2023obstacle,xiao2023positioning}. Transformer variants introduce global self-attention and alleviate spectral–spatial aliasing~\citep{Zhao2022RoadFormer,han2023uav}, yet most implementations still assume regularly sampled grids and require dense task-specific supervision. Transfer learning, self-supervised strategies~\citep{Naushad2021DeepTransfer,Alosaimi2023SelfSupervised,Scheibenreif2022SelfSupervised}, and diffusion-based generation~\citep{Sebaq2023RSDiff,Baghirli2023SatDM,Khanna2023DiffusionSat,Tang2024CRSDiff} partially mitigate data scarcity, but current systems either model complex landform geometries with inflexible kernels or augment training with synthetic imagery whose fidelity is insufficient for fine-grained, hydrology- and infrastructure-related land-cover types~\citep{DeVita2024DMGS}. We therefore concentrate on hydrology- and infrastructure-related categories—Beach, Bridge, Desert, Forest, Mountain, Pond, Port, and River in AID and Harbor, River, and Pond in SIRI-WHU—as a challenging, practically important testbed.

In response to these limitations, we propose LC4-DViT (Land-cover Creation for Land-cover Classification with Deformable Vision Transformer), an end-to-end framework that couples description-driven diffusion-based land-cover creation with a deformation-aware Vision Transformer. A prompt- and structure-aware generative pipeline combines Real-ESRGAN pre-enhancement, GPT-4o–generated class-specific descriptions, and ControlNet-guided Stable Diffusion to synthesize class-balanced, high-fidelity training images tailored to challenging land-cover types, while a DCNv4–ViT hybrid classifier captures both fine-scale geometry and global context. Our contributions are threefold: (i) a data-centric generative pipeline that directly addresses annotation scarcity and long-tailed imbalance; (ii) a deformation-aware Vision Transformer specifically designed for narrow, deformation-rich landforms such as rivers, ponds, ports, and bridges; and (iii) comprehensive evaluation on the AID subset and SIRI-WHU, including GPT-4o–based attention judging on cosine similarity heatmap, which shows that LC4-DViT outperforms vanilla ViT, ResNet-50, MobileNetV2, and FlashInternImage without increasing model size, while improving cross-dataset transferability.

\section{Materials and Methodology}
\label{sec:materials}

\subsection{Dataset}
Aerial Image Dataset (AID) is a large-scale, high-resolution benchmark for aerial scene classification, containing over ten thousand RGB images from Google Earth that closely approximate optical aerial photographs~\citep{Xia2017AID}. We use eight hydrology- and infrastructure-related classes—Beach, Bridge, Desert, Forest, Mountain, Pond, Port, and River—totaling 2,860 images: Beach and River capture dynamic shorelines and fluvial corridors, Pond represents still-water bodies, Port and Bridge model human–water interfaces important for urban water management, and Desert/Forest/Mountain provide geomorphic and land-cover contrasts relevant to runoff, watershed connectivity, and water availability. We first perform an 8:1:1 class-stratified split into training/validation/test; super-resolution and text-guided diffusion augmentation are applied only to the training images, producing one super-resolved and two synthetic images per original sample, while the test set remains non-augmented with 537 original images. After augmentation, we form new training and validation sets from the remaining original and synthetic images, yielding 7,552 training, 1,914 validation, and 537 test images (10,003 total). To assess cross-dataset generalization, we further evaluate on SIRI-WHU~\citep{Zhu2016BoVW} using the Harbor, River, and Pond classes (600 images), split per class with an 8:1:1 ratio into training/validation/test, and no super-resolution or diffusion-based augmentation is used (original images only).

\subsection{LC4-DViT Framework}

\begin{figure}[!t]
    \centering
    \includegraphics[width=\linewidth]{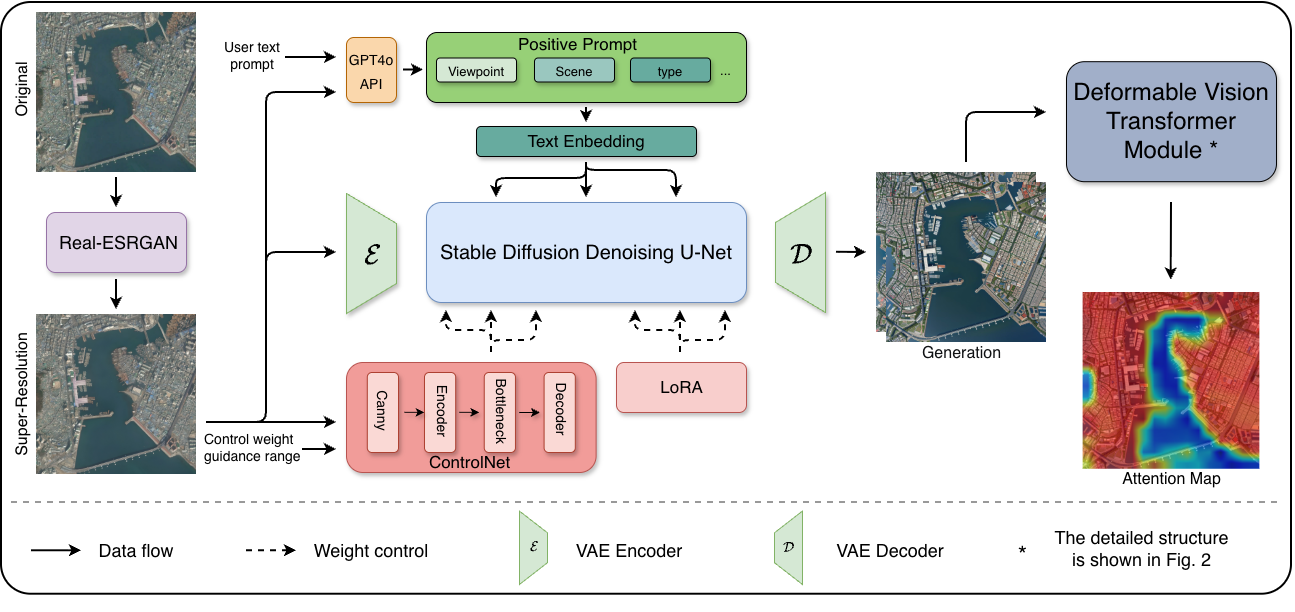}
    \caption{LC4-DViT Pipeline: Super-Resolution, GPT-4o/ControlNet Diffusion Augmentation and DViT Classification}
    \label{fig:LVC2-DViT}
\end{figure}

The LC4-DViT framework enhances remote-sensing land-use classification by combining data-centric augmentation with a deformation-aware classifier. As illustrated in Fig.~\ref{fig:LVC2-DViT}, the pipeline comprises three modules: (1) Real-ESRGAN-based super-resolution to sharpen small hydrological structures; (2) a GPT-4o–assisted, ControlNet-guided Stable Diffusion pipeline that synthesizes class-balanced, high-fidelity training images; and (3) a Deformable Vision Transformer (DViT) that couples a DCNv4 backbone with a ViT encoder to jointly capture fine-scale geometry and global context.

To enhance small hydrological and infrastructure features, we apply super-resolution to all AID training images. We employ the RRDBNet model within the Real-ESRGAN framework~\citep{Wang2021RealESRGAN} with a $4\times$ upscaling factor, which improves local detail while maintaining global structure. The super-resolved tiles are then used both as inputs to GPT-4o for description generation and as structural references for the subsequent diffusion process. We adopt a GPT-4o–assisted Stable Diffusion pipeline to enlarge the training corpus with high-fidelity, class-balanced imagery. For each training image, GPT-4o generates a Stable-Diffusion-style prompt that describes the scene content (viewpoint, composition, land-cover type). These prompts condition a domain-adapted Stable Diffusion model~\citep{Rombach2022}, fine-tuned on landscape imagery and equipped with a photo-realism LoRA, while negative prompts suppress artistic or cartoon-like textures.

To preserve spatial structure, we use ControlNet~\citep{Zhang2023} with Canny edge maps extracted from the super-resolved images as auxiliary conditions. This constrains the diffusion process to respect the geometry of rivers, bridges, beaches, and other salient objects, while allowing appearance-level diversification in surrounding regions. For each original training tile, we generate two synthetic images; combined with the super-resolved version, this yields a structurally consistent but visually diverse training set for LC4-DViT.

\begin{figure}[!t]
    \centering
    \includegraphics[width=0.95\linewidth]{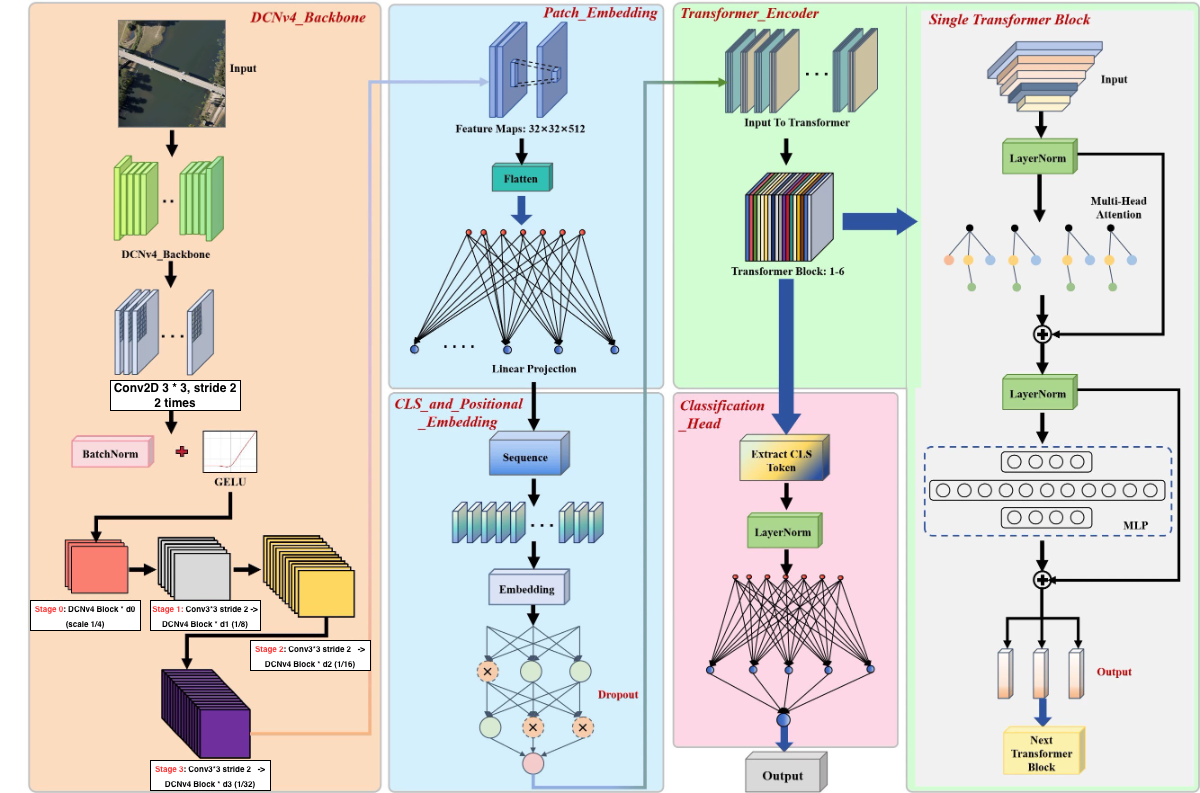}
    \caption{Overall architecture of the proposed Deformable Vision Transformer (DViT).}
    \label{fig:dvit}
\end{figure}

Finally, the DViT classifier (Fig.~\ref{fig:dvit}) is a hybrid architecture that combines a DCNv4-based hierarchical backbone with a ViT-style Transformer encoder. A lightweight convolutional stem and four DCNv4 stages first produce multi-scale, deformation-aware feature maps at progressively reduced resolutions; the final-stage feature map is then flattened and linearly projected into patch embeddings, augmented with a learnable classification (CLS) token and positional encodings, and fed into a Transformer encoder with standard self-attention and feedforward blocks. The CLS representation is passed through a linear head to predict the land-cover class. By delegating local, geometry-aware feature extraction to DCNv4 and global context modeling to the Transformer, DViT is well suited to elongated shorelines, meandering rivers, small ponds, ports, and bridges in high-resolution remote-sensing imagery.

To contextualize the performance of LC4-DViT, we fine-tune four representative backbones on the same training splits: 1.ResNet-50~\citep{he2016deep}: a widely used residual CNN with bottleneck blocks. 2.MobileNetV2~\citep{sandler2018mobilenetv2}: a lightweight CNN with inverted residual and depthwise separable convolutions. 3.FlashInternImage: an InternImage-style backbone built on DCNv4~\citep{xiong2024efficient,wang2022internimage} for deformable convolutional feature extraction. 4.Vision Transformer (ViT)~\citep{Dosovitskiy2020ViT}: a patch-based Transformer encoder with a classification token. All baselines share the same input resolution, optimization settings, and training protocol as DViT, enabling fair comparison of backbone design and data augmentation strategies.

\hspace{-3em}

\section{Results}
\label{sec:results}

\subsection{Environment Setup}

All experiments are implemented in Python~3.12 with PyTorch on NVIDIA RTX~4090 GPUs. To reduce randomness, each model is trained with ten different random seeds and we report the average of all metrics.

\subsection{Evaluation of Diffusion Framework}

To assess the image quality of our generative pipeline, we use the Kernel Inception Distance (KID), which measures the distributional similarity between features of generated and real images (lower is better)~\citep{Binkowski2018MMD}. As shown in Table~\ref{tab:kid_pid_metrics}, our LC4-DViT pipeline attains a KID of 12.10 on AID, substantially better than UniControl on the CNSATMAP dataset (20.42)~\citep{Qin2023UniControl,Dong2025EarthMapper}. This indicates that the synthesized images are reasonably aligned with the real data distribution and are of sufficient quality for downstream land-cover classification.

\begin{table}[!t]
  \centering
  \caption{Image quality evaluation using KID metrics.}
  \label{tab:kid_pid_metrics}
  \begin{tabular}{lcc}
    \toprule
    \textbf{Method} & \textbf{Dataset} & \textbf{KID} $\downarrow$  \\
    \midrule
    LC4-DViT (ours) & AID       & \textbf{12.10} \\
    UniControl & CNSATMAP & 20.42    \\
    \bottomrule
  \end{tabular}
\end{table}

\subsection{Evaluation of Model Performance}

We evaluate all backbones (ResNet50, MobileNetV2, ViT, FlashInternImage, and DViT) on AID using overall accuracy (OA), mean accuracy (mAcc), Cohen's Kappa, and macro precision, recall, and F1-score, following standard definitions~\citep{Hu2020UDA,Powers2020Evaluation}. All models are trained for 30 epochs with AdamW~\citep{Loshchilov2017} (learning rate $1\times10^{-4}$, batch size 16) and cross-entropy loss, using $512\times512$ inputs normalized with ImageNet statistics.

\begin{figure}[!t]
  \centering
  \includegraphics[width=0.85\linewidth]{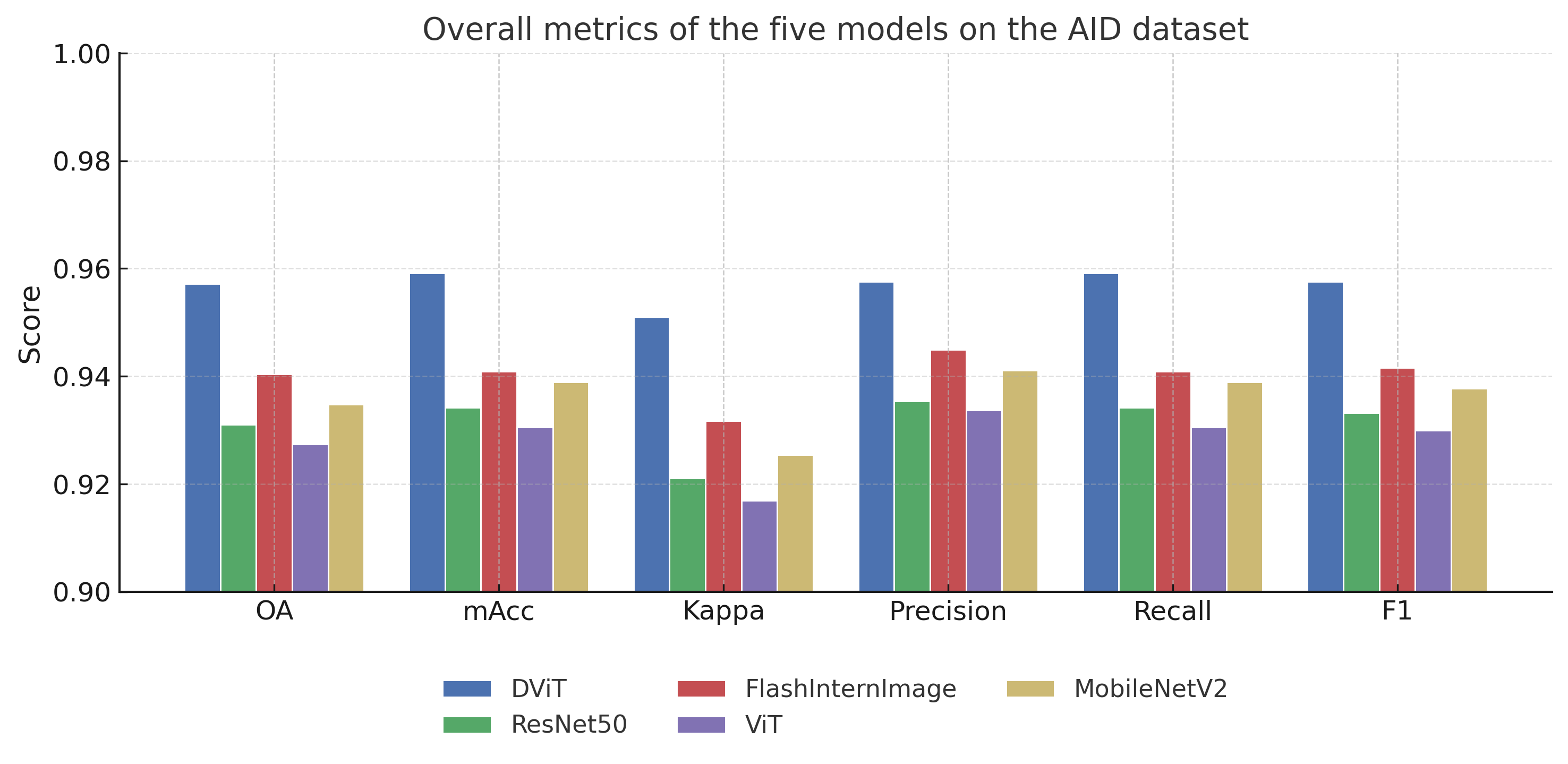}
  \caption{Overall metrics of the five models on the AID dataset.}
  \label{fig:overall_metrics}
\end{figure}

Figure~\ref{fig:overall_metrics} summarizes the overall performance of the five backbones on AID. DViT clearly dominates all competitors, achieving the highest OA (0.9572), mAcc (0.9592), Kappa (0.9510), and macro F1-score (0.9576), while keeping model size comparable to the other backbones. Compared with the strongest CNN baseline and the vanilla ViT, DViT yields gains of more than two percentage points in OA and mAcc, indicating a more balanced and robust classifier, particularly beneficial for hydrology- and infrastructure-related land-cover classes.

\subsection{Ablation Studies}

\begin{table}[!t]
  \centering
  \caption{Ablation results for Real-ESRGAN-based super-resolution (SupreR) and diffusion-based augmentation (Diffusion) on AID.}
  \label{tab:dvit_ablation}
  \resizebox{\columnwidth}{!}{%
  \begin{tabular}{lccccccccc}
    \toprule
    \textbf{Model} & \textbf{SupreR} & \textbf{Diffusion} &
    \textbf{Overall Accuracy} & \textbf{Mean Accuracy} &
    \textbf{Cohen-Kappa} & \textbf{Precision (macro)} &
    \textbf{Recall (macro)} & \textbf{F1-score (macro)} \\
    \midrule
    \multirow{4}{*}{DViT}
      &  &   
      & 0.9162 & 0.9185 & 0.9041 & 0.9213 & 0.9185 & 0.9183 \\
      & \checkmark & 
      & 0.9311 & 0.9341 & 0.9211 & 0.9326 & 0.9341 & 0.9326 \\
      &  & \checkmark
      & 0.9367 & 0.9383 & 0.9275 & 0.9386 & 0.9383 & 0.9380 \\
      & \checkmark & \checkmark
      & \textbf{0.9572} & \textbf{0.9592} & \textbf{0.9510} &
        \textbf{0.9576} & \textbf{0.9592} & \textbf{0.9576} \\
    \bottomrule
  \end{tabular}
  }
\end{table}

Table~\ref{tab:dvit_ablation} quantifies the effect of Real-ESRGAN super-resolution (SupreR) and diffusion-based augmentation (Diffusion). Starting from a baseline DViT trained only on the original images (OA 0.9162, Kappa 0.9041), enabling SupreR or Diffusion alone yields clear gains, with diffusion providing the larger improvement. Combining both leads to the best performance (OA 0.9572, Kappa 0.9510, macro F1 0.9576), showing that super-resolution and generative augmentation are complementary: SupreR sharpens small structures, while diffusion increases class balance and geometric diversity.

\subsection{Evaluation of Model Attention via Heatmaps}

\begin{figure}[!t]
    \centering
    \includegraphics[width=0.8\linewidth]{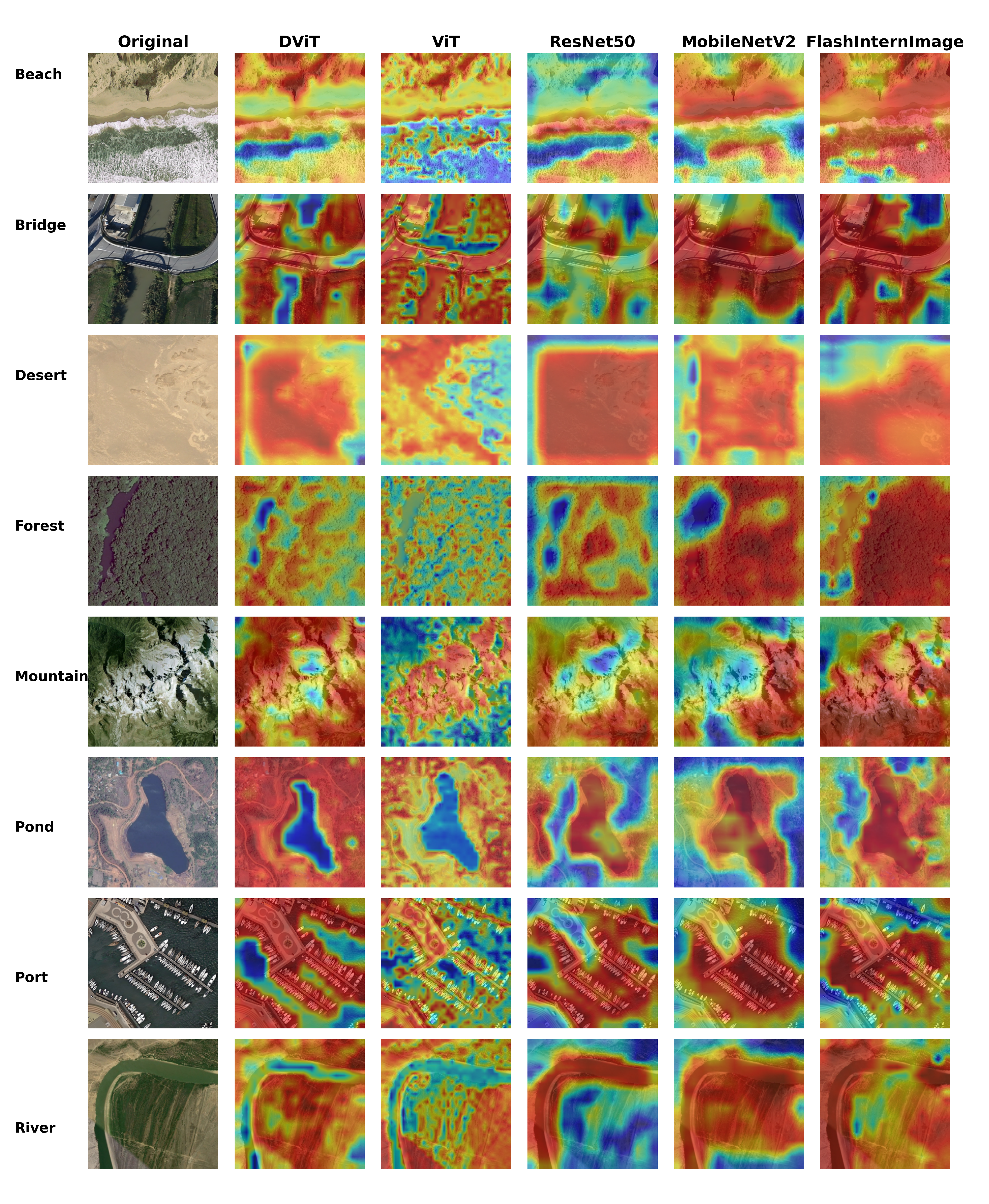}
    \caption{Class-activation heatmaps of different models for eight scene categories, which include the original scenes and the attention maps of the five backbones.}
    \label{fig:atten_heatmap}
\end{figure}

To analyze spatial attention, we visualize cosine-similarity heatmaps for all models (Fig.~\ref{fig:atten_heatmap}). Across categories, all backbones focus on semantically meaningful regions, but DViT produces the most compact and boundary-consistent responses, especially along water–land interfaces and slender structures such as bridges and port facilities, whereas the other models exhibit more diffuse or off-target attention. We further quantify alignment using a GPT-4o-based ``LLM judge'' that scores each heatmap from 0 to 3 according to how well high-activation regions follow class-discriminative landforms. As reported in Table~\ref{tab:llm_judge_avg}, DViT obtains the highest average score (2.625), confirming its superior spatial interpretability.

\begin{table}[t]
    \centering
    \caption{Average GPT-4o judge scores (0--3 scale) over eight sampled categories.}
    \label{tab:llm_judge_avg}
    \resizebox{\linewidth}{!}{%
    \begin{tabular}{lccccc}
        \toprule
            \textbf{Model} & \textbf{DViT} & \textbf{FlashIntern} & \textbf{ViT} & \textbf{ResNet50} & \textbf{MobileNetV2} \\
            \midrule
            \textbf{Average Score} & \textbf{2.625} & 2.5 & 2.25 & 2.0 & 2.5 \\
        \bottomrule
    \end{tabular}
    }%
\end{table}

\subsection{Cross-dataset Evaluation on SIRI-WHU}


To further validate generalization, we evaluate all models on a three-class SIRI-WHU subset (Harbor, Pond, River) without any augmentation. In this setting, DViT achieves the best performance (OA/mAcc = 0.9333, Kappa = 0.9000, macro Precision/Recall/F1 = 0.9406/0.9333/0.9316), exceeding the CNN baseline MobileNetV2 (OA/mAcc = 0.8833, Kappa = 0.8250, macro Precision/Recall/F1 = 0.8952/0.8833/0.8828) by 5.0 percentage points in OA/mAcc and 7.5 percentage points in Kappa. The remaining baselines show lower scores: ResNet-50 (OA/mAcc = 0.8667, Kappa = 0.8000, macro Precision/Recall/F1 = 0.8744/0.8667/0.8628), FlashInternImage (OA/mAcc = 0.8667, Kappa = 0.8000, macro Precision/Recall/F1 = 0.8808/0.8667/0.8656), and ViT (OA/mAcc = 0.8500, Kappa = 0.7750, macro Precision/Recall/F1 = 0.8541/0.8500/0.8510). These results indicate that the deformation-aware transformer transfers well across sensors and geographic domains, particularly for hydrology- and infrastructure-related land-cover types.

\section{Discussion}
\label{sec:discussion}

We propose LC4-DViT to examine how description-driven generative augmentation and a deformation-aware transformer jointly improve high-resolution land-cover mapping. Over ten runs on the selected AID subset, combining class-balanced diffusion sampling with DViT yields consistent gains: 0.9572 OA, 0.9592 mAcc, 0.9510 Kappa, and 0.9576 macro F1 (Fig.~\ref{fig:overall_metrics}). LC4-DViT surpasses the CNN baseline and vanilla ViT by more than 2 percentage points on all global metrics without increasing model capacity, with the largest improvements on hydrology- and infrastructure-related classes where objects are small and geometrically complex.

Ablations disentangle the contributions of data and architecture. Super-resolution and diffusion augmentation each improve DViT relative to training on the original imbalanced data, with diffusion providing the larger boost. The combination of Real-ESRGAN plus description-driven, ControlNet-guided diffusion—performs best, suggesting boundary sharpening and class-balanced, deformation-rich samples are complementary. Replacing the DCNv4 backbone with a standard ViT reduces accuracy and Kappa, underscoring the value of deformation-aware features for elongated shorelines, narrow rivers, ports, and bridges; GPT-4o–based assessment of cosine similarity further indicates stronger interpretability, with attention more consistently aligned to hydrologically meaningful structures across AID and SIRI-WHU. On a three-class SIRI-WHU subset (Harbor, Pond, River), DViT again achieves the best OA, macro F1, and Kappa, delivering roughly +5 points over the strongest CNN baseline in OA and mAcc, indicating good transfer across sensors and geographic domains.

\section{Conclusions}
\label{sec:conclusion}

In summary, LC4-DViT jointly addresses data scarcity and geometric complexity in land-cover classification. A GPT-4o–guided, Real-ESRGAN+Stable Diffusion pipeline produces class-balanced, high-fidelity training images, while a DCNv4–ViT hybrid captures both local geometric detail and global scene context. On AID and SIRI-WHU, LC4-DViT consistently outperforms strong CNN and transformer baselines and produces attention maps that external GPT-4o judges deem most consistent with class-defining landform structures. These results indicate that combining description-driven generative augmentation with deformation-aware transformers is a promising direction for high-resolution land-cover mapping and for improving recognition of hydrological and infrastructure classes central to flood-risk assessment, blue–green infrastructure planning, and broader environmental monitoring.

\small
\bibliographystyle{IEEEtranN}
\bibliography{references}

\end{document}